\pdfoutput=1

\documentclass[11pt]{article}

\usepackage[final]{acl}

\usepackage{times}
\usepackage{latexsym}

\usepackage[T2A]{fontenc}
\usepackage[T1]{fontenc}

\usepackage[utf8]{inputenc}

\usepackage{microtype}

\usepackage{inconsolata}

\usepackage{graphicx}

\usepackage{algorithm}
\usepackage{algpseudocode}
\usepackage{amsmath}
\usepackage{makecell}
\usepackage{multirow}
\usepackage{booktabs}

\title{Logos as a Well-Tempered Pre-train for Sign Language Recognition}

\author{
Ilya Ovodov, 
{\bf Petr Surovtsev},
{\bf Karina Kvanchiani},
\\
{\bf Alexander Kapitanov},
{\bf Alexander Nagaev}\\
SberAI\\
\small{
    \{iovodov, petr.surovcev, karinakvanciani, kapitanovalexander, sashanagaev1111\}@gmail.com
}
}

\begin{document}
\maketitle

\begin{abstract}

This paper examines two aspects of the isolated sign language recognition (ISLR) task.
First, although a certain number of datasets is available, the data for individual sign languages is limited.
It poses the challenge of cross-language ISLR model training, including transfer learning. Second, similar signs can have different semantic meanings. It leads to ambiguity in dataset labeling and raises the question of the best policy for annotating such signs.
To address these issues, this study presents Logos, a novel Russian Sign Language (RSL) dataset, the most extensive available ISLR dataset by the number of signers, one of the most extensive datasets in size and vocabulary, and the largest RSL dataset.
It is shown that a model, pre-trained on the Logos dataset can be used as a universal encoder for other language SLR tasks, including few-shot learning. We explore cross-language transfer learning approaches and find that joint training using multiple classification heads benefits accuracy for the target low-resource datasets the most. The key feature of the Logos dataset is explicitly annotated visually similar sign groups. We show that explicitly labeling visually similar signs improves trained model quality as a visual encoder for downstream tasks. Based on the proposed contributions, we outperform current state-of-the-art results for the WLASL dataset and get competitive results for the AUTSL dataset, with a single stream model processing solely RGB video.
The source code, dataset, and pre-trained models are publicly available.

\end{abstract}
\newcommand{\russian}[1]{\fontencoding{T2A}\fontfamily{cmr}\selectfont {#1}}
\begin{figure}[!ht]
  \centering
		\includegraphics[width=0.8\hsize]{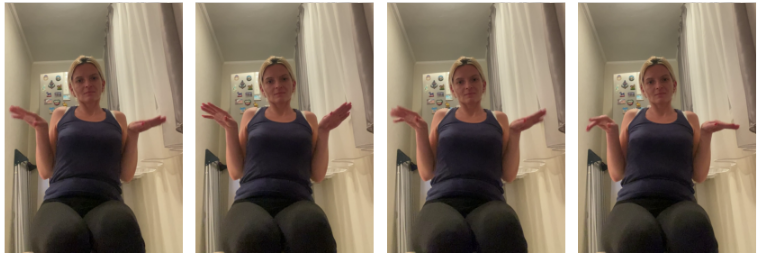} \\ (a) \russian{Летать} (to fly) %
		\vfill
		\includegraphics[width=0.8\hsize]{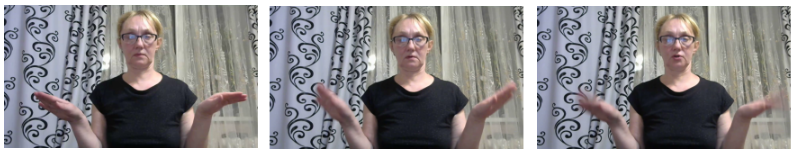} \\ (b) \russian{Крыло} (wing)
		\vfill
		\includegraphics[width=0.8\hsize]{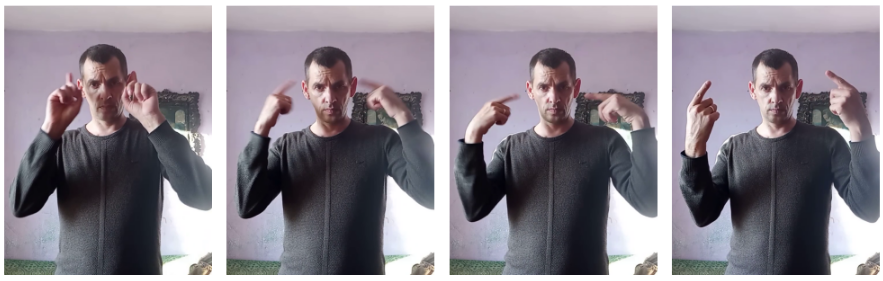} \\ (c) \russian{Овца} (sheep)
		\vfill
		\includegraphics[width=0.8\hsize]{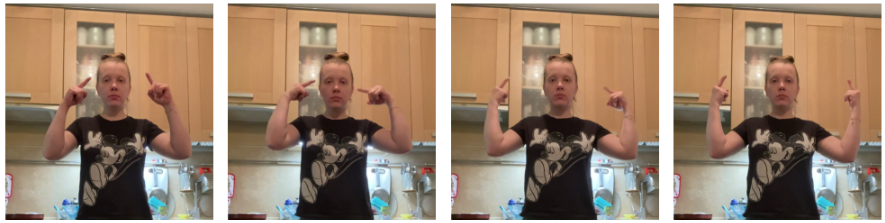} \\ (d) \russian{Разгневанный} (angry)
	\caption{Sample frames from Russian Sign Language dataset Logos: (a,b) and (c,d) are visually similar signs (VSSigns).}    
	\label{fig:01_sample}
\end{figure}

\section{Introduction}
\label{sec:intro}
Sign languages (SL) are visual-spatial signals for communication among deaf communities.
Although national sign languages are mostly associated with national spoken languages, they are distinct languages with their own grammar and vocabulary.
Primarily, signs are expressed by hand shape and motion (manual components of sign),
but also with a great aid of motion of mouth, head, eyes, and the body (non-manual components). 
The term ``gloss'' is used to refer to the word that signifies the sign. Generally, this word represents the sign's meaning. Still, one sign can be translated by several words and vice versa, so glosses should be considered just as word labels for signs.

The problem of computer sign language recognition and translation has a practical application with significant social impact because it can help deaf and hearing people communicate. On the other hand, it is a challenging scientific problem located at the junction of computer vision and natural language processing areas.

The presented work deals with the isolated sign language recognition (ISLR) problem, i.e., the classification of videos that contain only one sign each.
The ISLR task has not only independent significance but is also important for building a more practical continuous sign language translation (CSLT) solution \cite{chen2022simple,wei2023ImprovingContinuousSign,zuo2024towards}.

A significant obstacle to building SLR solutions is a shortage of training data~\cite{gokul2022AddressingResourceScarcity,papadimitriou2023SignLanguageRecognition}. While a number of annotated SL datasets exist, they represent different sign languages (Table~\ref{tab:datasets}), and dataset corpora for many individual languages are insufficient. It highlights the task of using cross-lingual data.
Some researchers utilize cross-lingual training either by combining two or more rather small datasets or using a multilingual dataset. The presented study examines the ability of a model trained on an extensive SL dataset of one language to serve as an encoder for SL tasks for other sign languages, and compares different approaches to it.

This paper presents an extensive Russian Sign Language (RSL) dataset, Logos, one of the largest existing sign language datasets in terms of volume and vocabulary size and the largest in terms of the number of signers.
We show that a model pre-trained on the Logos dataset can be successfully transferred to another language SLR tasks, including few-shot learning. The dataset size is critical, and the effect degrades if a smaller dataset is used for pre-train. 
Next, we compare transfer learning methods and find that simultaneous training with the large dataset using multiple classification heads for different languages benefits the target language SLR models the most, compared to other transfer learning methods.   

Another problem with SLR is that signs with similar handshapes and motions can have various semantic meanings. Such signs can be either strictly indistinguishable (the cases of polysemy or homonymy) or only distinguishable by their constituent non-manual features~\cite{zuo2023NaturalLanguageAssistedSign,hu2021GlobalLocalEnhancementNetwork}, see Figure \ref{fig:01_sample}. In the latter case, such signs can be viewed as minimal pairs for SL.
The difference between the individual signers' manners blurs the boundary between non-manual features and makes such signs practically indistinguishable out of context. For this reason, while some researchers consider such signs to be separate, others consider them to be the same~\cite{ebling2015factors}.
This paper calls such hardly distinguishable signs ``\textit{visually similar signs}'' (\textit{VSSigns}). This concept includes both polysemy and minimal pair cases. While they differ from a linguistic perspective, it makes sense to combine them for machine learning purposes. 

Different SL datasets have VSSigns annotated with either different or similar labels. 
To the best of our knowledge, no studies have examined the impact of the VSSigns annotation approach on resulting SLR models. We explore its effect in this work using the Logos dataset, which has both \textit{ungrouped gloss} and \textit{grouped VSSign} annotations. We find that VSSigns grouping benefits the SLR model.

The key contributions of this work are:

\begin{itemize}

    \item We present Logos, a new publicly available Russian Sign Language ISLR dataset.
    It is the most extensive available ISLR dataset by the number of signers and one of the largest datasets while also the largest RSL dataset in size and vocabulary.
    The dataset's key feature is an explicit annotation of visually similar sign (VSSign) groups.

    \item Using the Logos dataset, we show that explicitly grouping VSSign labels benefits trained model quality as a video encoder for downstream tasks like transfer learning to other sign languages.

    \item We show that a model, pre-trained on the proposed Logos dataset can be transferred to another language SLR tasks, including few-shot learning. We compare transfer learning methods and demonstrate that the method of cross-lingual multi-dataset co-training with multiple language-specific classification heads improves SL models for low-resource datasets the most, compared to the conventional ``pre-train and finetune'' method.

    \item Based on the described contributions, we obtain recognition accuracy for the American Sign Language dataset WLASL, superior to state-of-the-art (SOTA), with a single stream model processing solely RGB video.
    
\end{itemize}

The research was conducted in cooperation with the ``All-Russian Society of the Deaf'' (VOG). VOG experts and professional sign language interpreters participated at every stage of the Logos dataset creation. We also engaged deaf consultants in developing training strategies to apply considerations to specific solutions. Additionally, some of our researchers completed formal courses on RSL to enhance their knowledge in this domain.

The source code, dataset, and pre-trained models are publicly available\footnote{
https://github.com/ai-forever/logos
}.

\section{Related Works}
\label{sec:related}

\begin{table*}[t]
    \centering
        \scalebox{0.8}{
	\begin{tabular}{ l l l r r r l }
        \toprule
		\textbf{Dataset} & \textbf{Method} & \textbf{Language} & \textbf{ Samples } & \textbf{ Signers } & \textbf{Glosses} & \textbf{VSSigns} \\ 
        \midrule
		DEVISIGN-L \cite{wang2016IsolatedSignLanguage} & lab & Chinese (CSL) & 24,000 & 8 & 2,000 & -- \\
        SLR500 \cite{huang2019AttentionBased3DCNNsLargeVocabulary} & lab & Chinese (CSL) & 125,000 & 50 & 500 & -- \\

        MS-ASL \cite{joze2019MSASL} & web & American (ASL) & 25,513 & 222 & 1,000 & grouped \\

        SMILE \cite{ebling2018SMILESwissGerman} & lab & Swiss German & 9,000 & 30 & 100 & -- \\

        BosphorusSign22k \cite{ozdemir2020BosphorusSign22kSignLanguage} & lab & Turkish (TİD) & 22,542 & 6 & 744 & grouped \\

        AUTSL \cite{sincan2020AUTSLLargeScale} & lab & Turkish (TİD) & 38,336 & 43 & 226 & -- \\

        WLASL \cite{li2020WLASL} & web & American (ASL) & 21,083 & 119 & 2,000 & -- \\

        BSLDict \cite{momeni2020WatchReadLookup} & lab & British (BSL) & 14,210 & 28 & \textbf{9,283} & addressed\\

        K-RSL \cite{imashev2020k} & lab & Kazakh-Russian & 28,250 & 10 & 600 & addressed \\

        BSL-1K \cite{albanie2020BSL1K} & TV & British (BSL) & 273,000$^1$ & 40 & 1,064 & -- \\

        INCLUDE \cite{sridhar2020INCLUDE} & lab & Indian (ISL) & 4,292 & 7 & 263 & -- \\

        NMFs-CSL \cite{hu2021GlobalLocalEnhancementNetwork} & lab & Chinese (CSL) & 32,010 & 10 & 1,067 & addressed \\

        BOBSL \cite{albanie2021BBCOxfordBritishSign} & TV & British (BSL) & \textbf{452,000}$^1$ & 39 & 2,281 & -- \\

        GSL isol. \cite{adaloglou2021ComprehensiveStudyDeep} & lab & Greek (GSL) & 40,785 & 7 & 310 & grouped \\

        LSFB-ISOL \cite{fink2021LSFB} & lab & Fra/Bel & 47,600 & 100 & 395 & -- \\

        CISLR \cite{joshi2022CISLRCorpusIndian} & web & Indian (ISL) & 7,000 & 71 & 4,765 & -- \\

        LSA64 \cite{ronchetti2023LSA64ArgentinianSign} & lab & Argentinian & 3,200 & 10 & 64 & -- \\
     
        ASL Citizen \cite{desai2024ASLCitizenCommunitysourced} & crowd & American (ASL) & 83,399 & 52 & 2,731 & -- \\

        Slovo \cite{kapitanov2023SlovoRussianSign} & crowd & Russian (RSL) & 20,000 & 194 & 1,000 & -- \\

        FDMSE-ISL \cite{patra2024HierarchicalWindowedGraph} & lab & Indian (ISL) & 40,033 & 20 & 2,000 & -- \\

        MM-WLAuslan \cite{shen2024MMWLAuslanMultiViewMultiModal} & lab & Australian(Auslan) &
        282,000$^2$ & 76 &  3,215 & -- \\

        \midrule
        \textbf{Logos (Ours)} & crowd & Russian (RSL) & 200,000 & \textbf{381} & 2,863$^3$ & both \\
        \bottomrule
	\end{tabular}
    }
	\caption{Summary of existing ISLR datasets.
    \textit{Method} -- the collection method: laboratory, web scrapping, TV programs, crowdsourcing. 
    \textit{VSSigns} column shows if visually similar signs (VSSigns) were considered by the dataset authors:
    \textit{grouped} -- VSSigns groups have common labels;
    \textit{addressed} -- the authors adopt VSSigns presence in the dataset and propose some methods to tackle them at training time;
    "--" -- VSSigns presence is not discussed.\\
    $^1$~These datasets mostly have automatic annotations of isolated glosses.\\
    $^2$~Actually, the dataset contain 70730 samples recorderd from four views each.\\
    $^3$~This number of glosses is grouped into 2,004 VSSign labels. 
    }
	\label{tab:datasets}
\end{table*}

\subsection{Isolated Sign Language Recognition}
In recent years, a group of approaches for ISLR tasks rely on using RGB input data. Then, either 2D convolutional neural network (CNN) is applied to extract individual frames' features, followed by LSTM for the temporal component processing \cite{koller2019WeaklySupervisedLearning}, or the spatial and temporal components are simultaneously processed using 3D CNN \cite{papadimitriou2023SignLanguageRecognition,zuo2023NaturalLanguageAssistedSign,albanie2020BSL1K,huang2019AttentionBased3DCNNsLargeVocabulary,li2020WLASL,joze2019MSASL}.
After the proliferation of transformers, transformer-based image and video processing architectures were applied  \cite{kapitanov2023SlovoRussianSign,kvanchiani2024TrainingStrategiesIsolated}.
In addition to the RGB input, a depth map can be used \cite{jiang2021SkeletonAwareMultimodal,zuo2023NaturalLanguageAssistedSign}.

Another group of approaches utilizes pose (skeleton) keypoints and face landmarks generated by available frameworks \cite{hruz2022OneModelNot,jiang2021SkeletonAwareMultimodal,miah2023MultistreamGeneralGraphbased,papadimitriou2023SignLanguageRecognition,ryumin2023AudiovisualSpeechGesture}.
The skeleton keypoints can be represented as a sequence of heatmaps and processed similarly to video data \cite{zuo2023NaturalLanguageAssistedSign}. A series of methods build a graph based on physical skeleton connections and explore Graph Convolutional Networks (GCNs) \cite{hu2021SignBERT,hu2023signbert,patra2024HierarchicalWindowedGraph,zhao2023BESTBERTPretraining,jiang2021SkeletonAwareMultimodal}.

Most current SOTA SLR models are multi-stream and multi-modal and combine more than one of the methods listed above \cite{hruz2022OneModelNot,zuo2023NaturalLanguageAssistedSign,jiang2021SkeletonAwareMultimodal,miah2023MultistreamGeneralGraphbased,papadimitriou2023SignLanguageRecognition,ryumin2023AudiovisualSpeechGesture}.

Nevertheless, we focus our research on purely frontal RGB video, as this setup is most relevant for practical applications. Many possible real-world scenarios, such as educational tools for RSL and ASL learning, video conferencing, and public kiosk systems in environements like metro stations or airports, typically involve users facing a single RGB camera.

\subsection{ISLR Datasets}
The ISLR datasets differ in several aspects: language, collection method, size, vocabulary size, number of signers (see Table~\ref{tab:datasets}). 
The most common method of dataset collection is recording invited signers in laboratory conditions. However, this approach generally results in insufficient scene and signer variety, requiring the authors to record each video individually. Web scrapping of SL videos is rather effective and results in more diverse datasets. However, its serious problem is the absence of consent from the video owner and person represented in the video on the usage of the video as a part of the dataset. Albanie et al.~\shortcite{albanie2020BSL1K,albanie2021BBCOxfordBritishSign} prepared the British SL datasets using BBC TV programs with SL translation. The datasets are large but have limited scene variety and number of signers, and they mostly only have automatic annotation. Collecting video from SL experts using a web crowdsourcing platform has no problem with signers' consent and provides much more diverse footage. We have used this approach for our work.

Vocabulary size is critical for building a production-quality SLR model. We suppose that practically useful models must recognize over 1,000 glosses. Therefore, a massive number of video samples is needed to simultaneously satisfy both the requirements of a large number of glosses and of samples per gloss. Number of diverse signers is also important. As seen from Table~\ref{tab:datasets}, only a few datasets meet these requirements.

\subsection{The VSSigns Problem}
Some SL signs with different semantic meanings either can be considered as strictly indistinguishable (the cases of polysemy or homonymy) or can have similar handshapes and motions, but can only be distinguished by their constituent non-manual features~\cite{zuo2023NaturalLanguageAssistedSign,hu2021GlobalLocalEnhancementNetwork}, see Figure \ref{fig:01_sample}. The latter can be viewed as minimal pairs for SL. The border between polysemy and minimal pair signs can be blurred due to the individual signers' manners, making even signs with different non-manual features practically indistinguishable out of context. For this reason, while some researchers consider such signs to be separate, others consider them to be the same~\cite{ebling2015factors}. On the contrary, even the same sign can have distinct manual features depending on the context, i.e., a question vs. a statement~\cite{mukushev2020evaluation}. It results in ambiguity in the annotation of such signs. This paper calls such hardly distinguishable signs ``\textit{visually similar signs}'' (\textit{VSSigns}). Formally, we define them as signs that have different meanings but have the same manual component. The VSSign concept includes both polysemy and minimal pair cases. While they differ from a linguistic perspective, it makes sense to combine them for machine learning purposes.

There is no standard approach to annotating VSSigns: they can be annotated with either different or similar labels.
In this paper, we call it \textit{ungrouped gloss} and \textit{grouped VSSign} annotations. The datasets collected for the most common words of spoken language 
\cite{sincan2020AUTSLLargeScale,kapitanov2023SlovoRussianSign}, typical continuous phrases \cite{albanie2020BSL1K,albanie2021BBCOxfordBritishSign,adaloglou2021ComprehensiveStudyDeep}, or based on an SL dictionary \cite{patra2024HierarchicalWindowedGraph} primarily have different (\textit{ungrouped}) labels for similar signs. For instance, according to \cite{zuo2023NaturalLanguageAssistedSign}, among 2,000 classes of widely used WLASL dataset \cite{li2020WLASL}, 334 classes form groups of VSSigns. Additional efforts are needed to merge similar VSSigns and assign unique \textit{grouped VSSign} labels to them.

Among the reviewed datasets, three papers state that VSSigns were grouped. Three papers confirm the presence of ungrouped VSSigns in the presented datasets and propose some techniques to distinguish them (Table~\ref{tab:datasets}).
To improve VSSigns classification, Hu et al. \shortcite{hu2021GlobalLocalEnhancementNetwork} deform a feature map, stretching more informative areas to emphasize non-manual features. Zuo et al. \shortcite{zuo2023NaturalLanguageAssistedSign} propose label smoothing depending on their semantic difference and a common latent space for gloss embeddings and vision features to maximize the separability of confusing signs.
Other works do not mention any steps to handle VSSigns in the proposed datasets. To our knowledge, no research has examined the impact of VSSigns on the quality of the resulting encoder for downstream tasks. Such a study is one of the topics of this work, using transfer learning to another language as a downstream task example.

\subsection{Multi-Dataset Training}
Although researchers complain about insufficient SLR training data \cite{gokul2022AddressingResourceScarcity,papadimitriou2023SignLanguageRecognition}, the topic of cross-language dataset sharing is poorly exploited.
Gokul et al. \shortcite{gokul2022AddressingResourceScarcity} implemented a multilingual SLR model for 11 sign languages by simply translating the labels of all the languages into English. The authors themselves admit that their model of combining different languages is primitive and does not make progress for some datasets.
Tornay et al. \shortcite{tornay2020TowardsMultilingualSignLanguage} train a unified hand movement model using 3 different sign language resources.  Then, they optimize the classifier using the target sign language data.  However, their cross-lingual model falls short of the monolingual reference.
Yin et al. \shortcite{yin2022MLSLTMultilingualSign} propose the MLSLT translation network as a single model for multilingual translation. Their work is limited to their rather small datasets and doesn't address leveraging large SL datasets to improve the model quality.
SignCLIP~\cite{jiang2024SignCLIP} utilizes a multilingual sign language dictionary, SpreadTheSign.com, for training. It contains mainly a single sample per concept per language, and the SignCLIP authors translate all concepts into English, similar to~\cite{gokul2022AddressingResourceScarcity}, so knowledge transfer between different sign languages is not investigated.
Hu et al. \shortcite{hu2022CollaborativeMultilingualContinuous} introduced an additional shared module that learns knowledge from two languages. It improved accuracy for Chinese CSL-Daily \cite{zhou2021improving} and German Phoenix-14 \cite{koller2015ContinuousSignLanguage} datasets. 
Wei et al. \shortcite{wei2023ImprovingContinuousSign} also benefit from the joint using the same datasets by creating a gloss translation map based on the visual similarity of signs, rather than their meanings. Authors train the model for the German language using both datasets and replace gloss labels in Chinese videos with German labels using this map, treating Chinese signs as German. As shown below, this mapping method does not give optimal results (see Section \ref{sec:ablation:mapping}).
However, as far as we know, the ability of a model trained on an extensive SL dataset to serve as an encoder for SL tasks for other sign languages has not been explored enough. Such a study, along with the comparison of different approaches to it, is another subject of our work.
\section{Logos Dataset}
\label{sec:logos}

\subsection{Dataset Characteristics}
The Logos dataset contains 199,668 videos recorded by 381 signers (deaf individuals, professional interpreters, sign language teachers, and CODA/SODA\footnote{Child of Deaf Adult, Sibling of Deaf Adult/Deaf person}).
The total duration of the dataset video is 221.4 hours, with 104.7 hours representing the demonstration of signs themselves and the rest being fragments before and after the sign demonstration.
The dataset contains signs for 2,863 of the most commonly used lemmas in the Russian general vocabulary, combined into 2,004 grouped VSSign classes with 35 to 737 samples per class.

The Logos dataset includes the Slovo public dataset \cite{kapitanov2023SlovoRussianSign} with the renewed annotations, amended with VSSign classes. 

More details on the dataset's characteristics are provided in Appendix~\ref{app_subsec:dataset_char}.

\subsection{Gloss Selection}
The Logos vocabulary selection is based on the frequency list of the Russian language corpus\footnote{\url{http://dict.ruslang.ru/freq.php}}. We have (1) selected the top 3,000 lemmas, except for prepositions, conjunctions, particles, and interjections, (2) removed lemmas that present in the Slovo dataset, and (3) selected glosses as lemmas for which sample video present on the SpreadTheSign\footnote{\url{https://spreadthesign.com/ru.ru/search/}} sign language
dictionary website.
We added 1,863 new glosses, bringing the total in the Logos dataset to 2,863 glosses.

\subsection{Data Collection}
The Logos data collection pipeline includes signer selection, video collection, video validation, and sign time interval annotation stages that coincide with the ones of the Slovo\cite{kapitanov2023SlovoRussianSign} pipeline. See Appendix~\ref{app_subsec:dataset_collection} for details.

\subsection{VSSigns Grouping}
We grouped visually similar signs based solely on their manual components through two stages.

First, we trained a baseline model on the dataset with ungrouped glosses and processed 2,863 sign template videos with the model. Using confidence of prediction classes for the template videos, we identified the 10 most similar templates for each one.
Deaf experts compared each template video with its 10 counterparts and marked pairs that differ only in non-manual components.  Sign pairs matched by the majority of 5 experts were annotated as VSSigns.

Next, we applied three rounds of additional verification. 
In each round, the model was trained on the currently grouped labels.
Based on the classification results, we identified the most confusing class pairs and visually inspected misclassified samples.
If VSSign candidates were found, we consulted deaf experts and grouped the labels additionally.

\subsection{Train-test Split}
We aim to maintain an 80/20 ratio for the train and test data split applied to both the number of signers and the number of samples for each sign.
Given that the number of signs recorded by different signers differs, the dataset split confirming all these requirements hardly has a strict resolution.
We applied a dynamic programming algorithm to find the best approximation.
See Appendix~\ref{app_subsec:split}.

\section{Experiments setup}
\label{sec:setup}

\subsection{Datasets}
\label{sec:setup:datasets}

In addition to the proposed large-scale Logos dataset, we selected two widely used ISLR benchmarks as target datasets for transfer learning: the Turkish Sign Language (TİD) dataset AUTSL~\cite{sincan2020AUTSLLargeScale} and the American Sign Language (ASL) dataset WLASL~\cite{li2020WLASL}.
The WLASL dataset offers a large vocabulary but is relatively small in size, comprising an average of approximately 10 samples per class.
The AUTSL dataset provides more samples but includes a more limited vocabulary and fewer signers.
We also include MM-WLAuslan \cite{shen2024MMWLAuslanMultiViewMultiModal}, one of the most extensive publicly available ISLR datasets, as an additional pre-training baseline to compare against Logos.
Key characteristics of these datasets are summarized in Table~\ref{tab:datasets}. 

\subsection{Sign Language Recognition Pipeline}
\label{sec:setup:pipeline}

Our experimental setup is based on ~\cite{kvanchiani2024TrainingStrategiesIsolated}. The authors explore various training aspects to propose the optimal ISLR pipeline. They use MViTv2-S~\cite{li2022mvitv2} as a backbone, a fully connected (FC) layer for classification, a cross-entropy classification loss with label smoothing, and sign timeline boundary regression as an auxiliary task.
The backbone was initialized with Kinetics-400 pre-train.
The pipeline processes $32\times224\times224$ frame chains, randomly sampled from the input video with a step of 2 frames.
We implement an auxiliary boundary regression task as follows.
The sign's ground truth boundary timestamps are rescaled relative to the sampled clip: the clip length is set as 1, the clip start is set as 0 for the sign start point, and the clip end is set as 0 for the sign endpoint.
Alongside the classification heads, we add an extra FC layer with two output channels for the sign start and end regression. Its output and scaled ground truth values are mapped to $(-1,1)$ using the formula $y=2\sigma(x)-1$, where $\sigma(x)$ is the sigmoid function, to diminish the influence of sign boundaries that are outside the clip.
We use mean squared error loss to train this regression head. The total loss function is calculated as a weighted sum of the classification and regression losses: $L = L_{cls} + 2.5 L_{regr}$.
We evaluate the model using a top-1 instance-based accuracy metric: the ratio of the correctly classified samples to the total samples number.

\subsection{Multi-dataset Co-training Method}
\label{sec:setup:multi}

\begin{figure*}[!htb]
    \centering
    \includegraphics[width=0.85\linewidth]{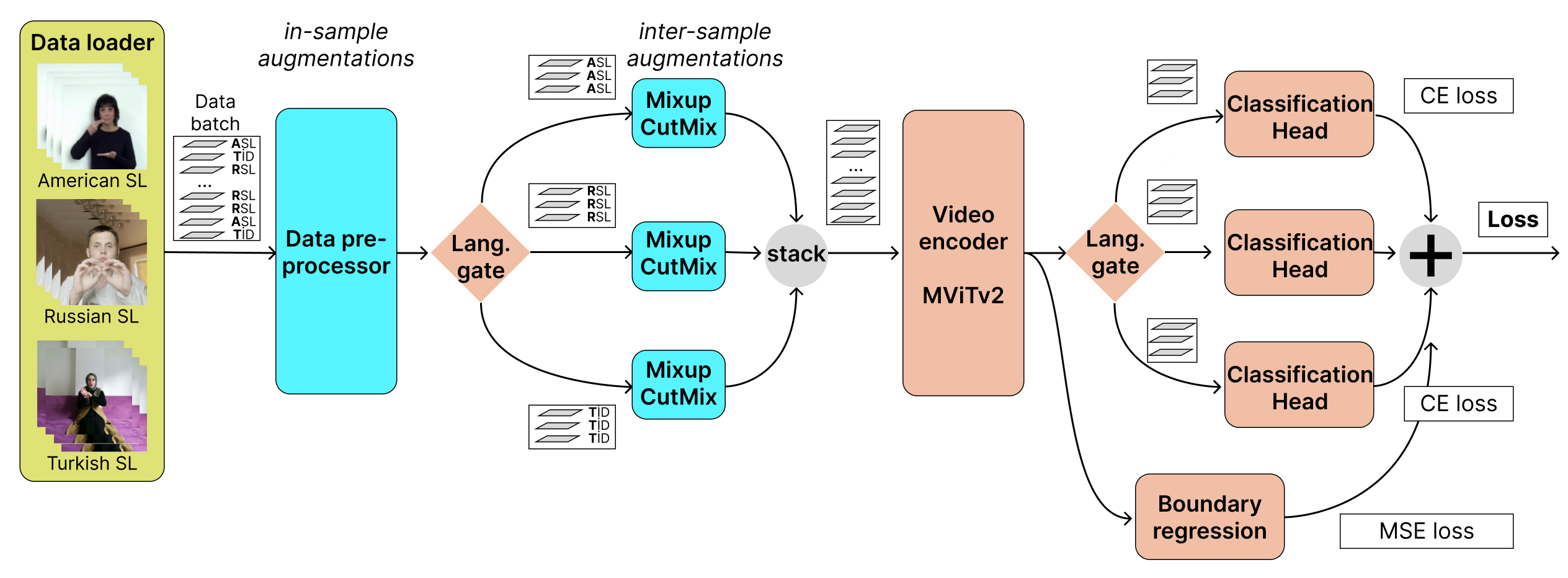}
    \caption{Multi-dataset co-training pipeline. Samples from different languages are processed as a united batch. Before the inter-sample augmentations and the language-specific classification heads, the language-specific gates split the batch into language-specific sub-batches.}
    \label{fig:multi-pipe}
\end{figure*}

Different national sign language datasets have their own label spaces with no common taxonomy. As a result, they cannot be directly combined for joint training.

In our pipeline (Figure~\ref{fig:multi-pipe}), we mark each sample with its language tag. During training, we form batches containing a mix of sign languages. After processing the mixed batch by the common visual encoder, we apply the 
language-specific gate, which splits the batch into language-specific sub-batches using the language tag and processes each sub-batch by the language-specific classification head. Loss functions from each classification head were weighted proportionally to the number of appropriate language samples in the mixed batch.

At the training stage, we use CutMix~\cite{yun2019CutMixRegularizationStrategy} and Mixup~\cite{zhang2018MixupEmpiricalRisk} inter-sample regularization strategies. They can not be applied to the mixed batch because labels of different languages cannot be mixed. We use the same
language-specific gate to split the mixed batch into language-specific sub-batches before applying these augmentations and then merge the resulting samples back into one batch.

\section{Experiment Results and Ablation Study}
\label{sec:ablation}

\subsection{Transfer learning experiments}
\label{sec:ablation:baseline}

\begin{table}[t]
    \centering
    \scalebox{0.78}{
	\begin{tabular}{ l c   c  c }
        \toprule
        \multirow{2}{*}{\textbf{Method}} & \multicolumn{3}{c}{\textbf{Top-1 accuracy}} \\
        \cmidrule{2-4} 
        & Logos & AUTSL & WLASL \\
        \midrule
    	Separate training (baseline) & 97.90 & 96.58 & 60.88 \\
        \midrule
        \multicolumn{4}{l}{\textbf{Transfer learning from Logos dataset:}} \\
	  Encoder is frozen        & -- & 97.25 & 62.44 \\
        Encoder is being trained & -- & 97.73 & 65.57 \\
        \midrule
        \multicolumn{4}{l}{\textbf{Multi-dataset co-training with Logos dataset:}} \\
        Logos + AUTSL          & 97.92 & \textbf{97.83} & -- \\
        Logos + WLASL          & \textbf{97.93} & --   & 65.74 \\
        Logos + AUTSL + WLASL & 97.92 & 97.81 & \textbf{66.82} \\
        \bottomrule
	\end{tabular}
    }
	\caption{Baseline, transfer learning, and multi-dataset co-training with the Logos dataset. Transfer learning and Multi-dataset co-training experiments use the encoder, initialized from the Logos pre-train.
    }
	\label{tab:transfer}
    
\end{table}

\begin{table}[t]
    \centering
    \scalebox{0.78}{
	\begin{tabular}{ l c  c }
    \toprule
    \multirow{2}{*}{\textbf{Method}} & \multicolumn{2}{c}{\textbf{Top-1 accuracy}} \\
    \cmidrule{2-3} 
    & AUTSL & WLASL \\
    \midrule
    Full dataset (baseline) & 97.25 & 62.44 \\
    \midrule
    10-shot (10 samples per class) & 90.16 & 61.12 \\
    3-shot (3 samples per class) & 83.99 & 54.10 \\
    one-shot (1 sample per class) & 82.44 & 37.07 \\
    \bottomrule
    \end{tabular}}
	\caption{Few-shot and one-shot transfer learning with frozen Logos pre-trained encoder.}
	\label{tab:few_shot}
\end{table}

The presented extensive Logos dataset was used as a pre-train for transfer learning tasks.
We used relatively small AUTSL and WLASL datasets as modeling examples of low-resource datasets for transfer learning. These datasets were also selected because benchmark results are available for comparison. 
Additionally, we created reduced versions of each dataset, limited to 10, 3, and 1 sample per class, for more challenging low-data experiments.

First, we trained the separate baseline models on the Logos, AUTSL, and WLASL datasets using the same setup (Section~\ref{sec:setup}).
Then, we examined the applicability of the Logos pre-trained model for transfer learning to smaller AUTSL and WLASL datasets. With the model backbone initialized from the Logos pre-train, we evaluated two transfer learning strategies: (a) training all model weights and (b) freezing the pre-trained encoder and training only the classification head. The Logos pre-train substantially improves the model accuracy compared to training from scratch (Table~\ref{tab:transfer}).

Next, we explored the potential of a Logos pre-trained encoder for few-shot learning on other sign languages. We limited train sets of AUTSL and WLASL datasets to the randomly selected 10, 3, and 1 samples per class. Then, we applied transfer learning with a frozen encoder to these truncated datasets. The test part of the datasets was left intact.
Although truncated datasets produce worse models, training even on 1 sample per class still keeps the models working, at least for the AUTSL dataset, which has a smaller vocabulary (Table~\ref{tab:few_shot}).

These experiments demonstrate the possibility of transfer learning from the extensive Logos dataset to other sign languages with only a limited amount of training data. 

\subsection{Cross-lingual Multi-dataset Co-training}
\label{sec:ablation:co-training}

We investigated the described multi-dataset co-training method using the pairs Logos and AUTSL, Logos and WLASL, and all three datasets combined.  The encoder and the Logos classifier were initialized from the Logos baseline model for all experiments.

A single model, produced by a multi-dataset co-training, far surpasses the accuracy of the models, separately trained on low-resource datasets from scratch, 
and also surpasses individual models trained using conventional transfer learning (Table~\ref{tab:transfer}). Moreover, results for the WLASL dataset are far above existing SOTA metrics\footnote{according to \url{https://paperswithcode.com/} and other papers referring to the datasets in question}, see Table~\ref{tab:SOTA}.
As for the AUTSL dataset, note that all leading models use ensembling, pose recognition, and depth maps (or some of the above). 
In contrast, our model uses a single stream that takes only RGB input.

\begin{table}[t]
    \centering
    \scalebox{0.78}{
	\begin{tabular}{ l  c  c  }
		\toprule
		\multirow{2}{*}{\textbf{Model}} & \multicolumn{2}{c}{\textbf{Top-1 accuracy}} \\
        \cmidrule{2-3} 
        & AUTSL & WLASL \\
        \midrule
        BSL-1K \cite{albanie2020BSL1K}  & -- & 46.9 \\
        SignBERT \cite{hu2021SignBERT} & -- & 54.7 \\
        SAM-SLR \cite{jiang2021SkeletonAwareMultimodal} & 98.5 & 58.7 \\
        One Model is Not Enough$^1$ & 96.4 & --\\
        ZBEST \cite{zhao2023BESTBERTPretraining} & -- & 54.6 \\
        SignBERT+ \cite{hu2023signbert} & -- & 55.6 \\
        NLA-SLR \cite{zuo2023NaturalLanguageAssistedSign} & -- & 61.3 \\
        SL-GDN \cite{miah2023MultistreamGeneralGraphbased} & 96.5 & --\\
        ST-GCN$^2$ & 96.7 & --\\
        Audio-visual ... \cite{ryumin2023AudiovisualSpeechGesture} & \textbf{98.6} & -- \\
        HWGAT \cite{patra2024HierarchicalWindowedGraph} & 95.8 & 48.5 \\
        StepNet \cite{shen2024StepNetSpatialtemporalPartaware} & -- & 61.2 \\
        Uni-Sign \cite{li2025UniSignUnifiedSign} & -- & \textbf{63.5} \\
        
        \midrule
	  MViTv2 (our baseline) & 96.58 & 60.88 \\
        Multi-dataset with MM-WLAuslan & 97.43 & 64.59 \\
        \textbf{Multi-dataset with Logos (ours)} & \textbf{97.81} & \textbf{66.82} \\
        \bottomrule
	\end{tabular}
    } \\ 

	\caption{Our results compared with SOTA results for the AUTSL and WLASL datasets.\\
    $^1$~\cite{hruz2022OneModelNot}\\
    $^2$~\cite{papadimitriou2023SignLanguageRecognition} }
	\label{tab:SOTA}
\end{table}

\subsection{The Encoder Generalization Ability Check}
\label{sec:ablation:mapping}

We examined the hypothesis that an encoder pre-trained on the Logos dataset does not produce universal sign features but can only recognize the signs of the pre-train language. 
When applied to another language, the model maps these signs to the most similar target language signs, as in the approach of \cite{wei2023ImprovingContinuousSign}. To emulate this hypothesis, we processed the WLASL train set with the Logos pre-trained model and built the map by associating the assigned Logos labels with the most frequent WLASL ground truth labels. Then, we applied the same model to the WLASL test set and substituted the resulting Logos labels with WLASL labels using the map instead of training a target language classification head. We repeated the same experiment with the AUTSL dataset.   

The results in Table~\ref{tab:table_dict} show that although this label mapping method works, it is significantly inferior to the trained classifier for the Logos pre-trained encoder. It confirms that the Logos pre-trained encoder produces universal sign embeddings that can encode new, unseen signs from another language.

\begin{table}[t]
    \centering
    \scalebox{0.78}{
	\begin{tabular}{ l c c }
		\toprule
		\multirow{2}{*}{\textbf{Method}} & \multicolumn{2}{c}{\textbf{Top-1 accuracy}} \\
        \cmidrule{2-3}
        & AUTSL & WLASL \\
        \midrule
		Transfer learning  & \textbf{97.25} & \textbf{62.44} \\
	  Map labels to target language & 65.78 & 23.63 \\
        \bottomrule
	\end{tabular}}
	\caption{Transfer learning with frozen encoder compared to label mapping from Logos to other language datasets.}
	\label{tab:table_dict}
\end{table}

\subsection{The Importance of the Pre-train Dataset}
\label{sec:ablation:size}
\begin{table}[t]
    \centering
    \scalebox{0.77}{
	\begin{tabular}{ l  c  c  c  c }
		\toprule
        \multirow{3}*{\makecell{
            \textbf{Pre-train}\\
            \textbf{dataset}
            }}        
        & \multicolumn{4}{c}{\textbf{Top-1 accuracy}} \\
            \cmidrule{2-5}
            & AUTSL& AUTSL, & WLASL & WLASL, \\
            &      & 3-shot &       & 3-shot \\
        \midrule
		Logos & \textbf{97.25} & \textbf{83.99} & \textbf{62.44} & \textbf{54.10} \\       
	  AUTSL  & -- & -- & 28.46 & 18.76 \\
	  WLASL  & 93.16 &  67.90 & -- & -- \\
        MM-WLAuslan & 93.64 & 69.79 & 45.21 & 33.01 \\
        \bottomrule
	\end{tabular}%
    }
	\caption{
    The importance of the pre-train dataset for cross-language transfer learning. Results for both whole and truncated versions of the AUTSL and WLASL datasets using pre-training on the Logos, WLASL, AUTSL, and MM-WLAuslan datasets.
    }
	\label{tab:table_small}
\end{table}

Table~\ref{tab:table_small} demonstrates that extensive dataset size is critical for training a powerful encoder for cross-language transfer learning. We repeated transfer learning experiments using pre-train on smaller AUTSL and WLASL datasets. One can see that the resulting accuracy degrades substantially compared to Logos pre-train.

We also compared the impact of the Logos dataset and the large-scale MM-WLAuslan dataset on multi-dataset co-training and transfer learning (Tables~\ref{tab:SOTA},\ref{tab:table_small}). The results obtained with Logos are substantially better than those with MM-WLAuslan for both tasks.

\subsection{The Effect of VSSigns Grouping}
\label{sec:ablation:gloss}

\begin{table}[t]
    \centering
    \scalebox{0.78}{
	\begin{tabular}{c c c c c }
		\toprule
        \multicolumn{2}{c}{\textbf{VSSigns}}
             & \multicolumn{3}{c}{\textbf{Top-1 accuracy}} \\
        \cmidrule(r){1-2} \cmidrule(l){3-5}
        \textbf{train} & \textbf{test} & Whole & non-VSSigns & VSSigns \\
        
        \midrule
		Yes & Yes & \textbf{97.90} & \textbf{97.49} & \textbf{98.33} \\
	  No & Yes  & 97.44 & 97.10 & 97.79 \\
	  No & No   & 87.02 & 97.10 & 76.51 \\
        \bottomrule
    \end{tabular}
    }
    \caption{Comparison of training using grouped VSSigns annotation (baseline) and annotation without grouping.}
	\label{tab:table_word_vs_gloss}
\end{table}

\begin{table}[t]
	\begin{center}
    \scalebox{0.78}{
	\begin{tabular}{ c c c c c }
		\toprule
		\multirow{3}{*}{\makecell{\textbf{Logos}\\\textbf{pre-train}\\\textbf{on VSSigns}}} & \multicolumn{4}{c}{\textbf{Top-1 accuracy}} \\
        \cmidrule{2-5}
            & AUTSL & AUTSL, & WLASL & WLASL, \\
            &       & 3-shot &       & 3-shot \\
        \midrule
		Yes & \textbf{97.25} & \textbf{83.99} & \textbf{62.44} & \textbf{54.10} \\
    	No & 96.79 & 82.38 & 60.74 & 51.60 \\
        \bottomrule
    \end{tabular}%
    }
    \end{center}    
	\caption{The effect of VSSigns grouping on transfer learning. Results for WLASL and AUTSL (whole and truncated to 3 samples per class) trained from Logos pre-train on grouped VSSigns annotation (baseline) and annotation without grouping.}
	\label{tab:table_words_transfer}
\end{table}

We investigated the contribution of our approach with grouping labels of visually similar signs in obtaining a high-quality encoder. We trained the classifier on the Logos dataset, using unique pairs of ungrouped and grouped labels as classes.
It formed 2,863 ungrouped gloss classes instead of 2,004 grouped VSSign classes in the baseline Logos annotation. Each ungrouped label has a unique associated grouped label, so the model, trained on the ungrouped labels, can be evaluated on grouped labels.

Table~\ref{tab:table_word_vs_gloss} shows the accuracy for models trained on 2,004 VSSign and 2,863 ungrouped classes. Quite predictably, the last model yields lower accuracy, primarily due to confusion of VSSigns. However, it is essential that it achieves lower accuracy on grouped VSSigns classes (the classes on which the 1st model was trained). Notably, the degradation is observed even for signs that are not VSSigns whose labels do not differ in all cases..
Furthermore, Table~\ref{tab:table_words_transfer} shows that VSSigns grouping results in more effective transfer learning to other sign languages.

\section{Conclusion}
\label{sec:conclusions}

The paper examines two aspects of the isolated sign language recognition (ISLR) task: cross-language SL model training, including transfer learning, and approaches to handling visually similar signs (VSSigns).
To explore these issues, this work presents Logos, a new publicly available Russian Sign Language dataset, the most extensive ISLR dataset by the number of signers and one of the largest available datasets while also the largest RSL dataset in size and vocabulary.
It is shown that a model, pre-trained on the Logos dataset can be used as a universal encoder for other language SLR tasks, including few-shot learning.
The cross-language transfer learning methods are evaluated, and it is demonstrated that the method of multi-dataset co-training with multiple language-specific classification heads improves SL models for low-resource datasets the most, compared to the conventional ``pre-train and finetune'' method.
The key feature of the Logos dataset is the explicit annotation of visually similar sign groups. With its use, we show that explicitly grouping VSSign labels benefits trained model quality as a video encoder for downstream tasks, such as transfer learning to other sign languages.
Based on the proposed contributions, we outperform current state-of-the-art results for the WLASL dataset and get competitive results for the AUTSL dataset, with a single stream model processing solely RGB video.

\section*{Limitations}

This work is limited by using the MViT baseline architecture and focusing on cross-language transfer learning in ISLR as the downstream task. To generalize the conclusions, further research is needed involving diverse model architectures, low-resource target datasets, and a broader range of downstream tasks, including continuous sign language recognition. We consider this a promising direction for future work.

The Logos dataset reflects the demographics of the participants involved in its collection, resulting in an unbalanced distribution in terms of age and gender. Additionally, the dataset focuses exclusively on RSL, which limits its direct applicability to more diverse settings. Nonetheless, our findings suggest it can still be effectively leveraged in broader sign language recognition tasks.
\section*{Ethical Statement}

Legal and ethical aspects were reviewed and approved by our institution's legal team.
All crowdworkers provided informed consent, authorizing the processing and publication of the collected data.
Informed consent was provided via a textual form on the crowdsourcing platform. Since all participants were fluent in written Russian, no interpreter translation was required.
To save contributors' privacy, we use anonymized user hash IDs.
We do not restrict the participation of signers under 18, provided parental consent was obtained during registration, in compliance with the Civil Code of the Russian Federation.
Participation was voluntary.
Compensation for completed tasks was aligned with the average salary of a sign language interpreter, proportionate to the time invested.
We have verified that the Slovo dataset, incorporated into Logos, adheres to these ethical standards.
The dataset is made available exclusively for research purposes.
Nonetheless, we acknowledge the potential misuse, such as identifying individuals or enabling large-scale surveillance.

\bibliography{main}

\appendix

\section{Logos Dataset}
\label{sec:logos_appendix}

This appendix contains more detailed characteristics of the Logos dataset and additional technical details for some production steps.

\subsection{Dataset Characteristics}
\label{app_subsec:dataset_char}
\begin{figure*}[ht]
    \centering
    \includegraphics[width=0.90\linewidth]{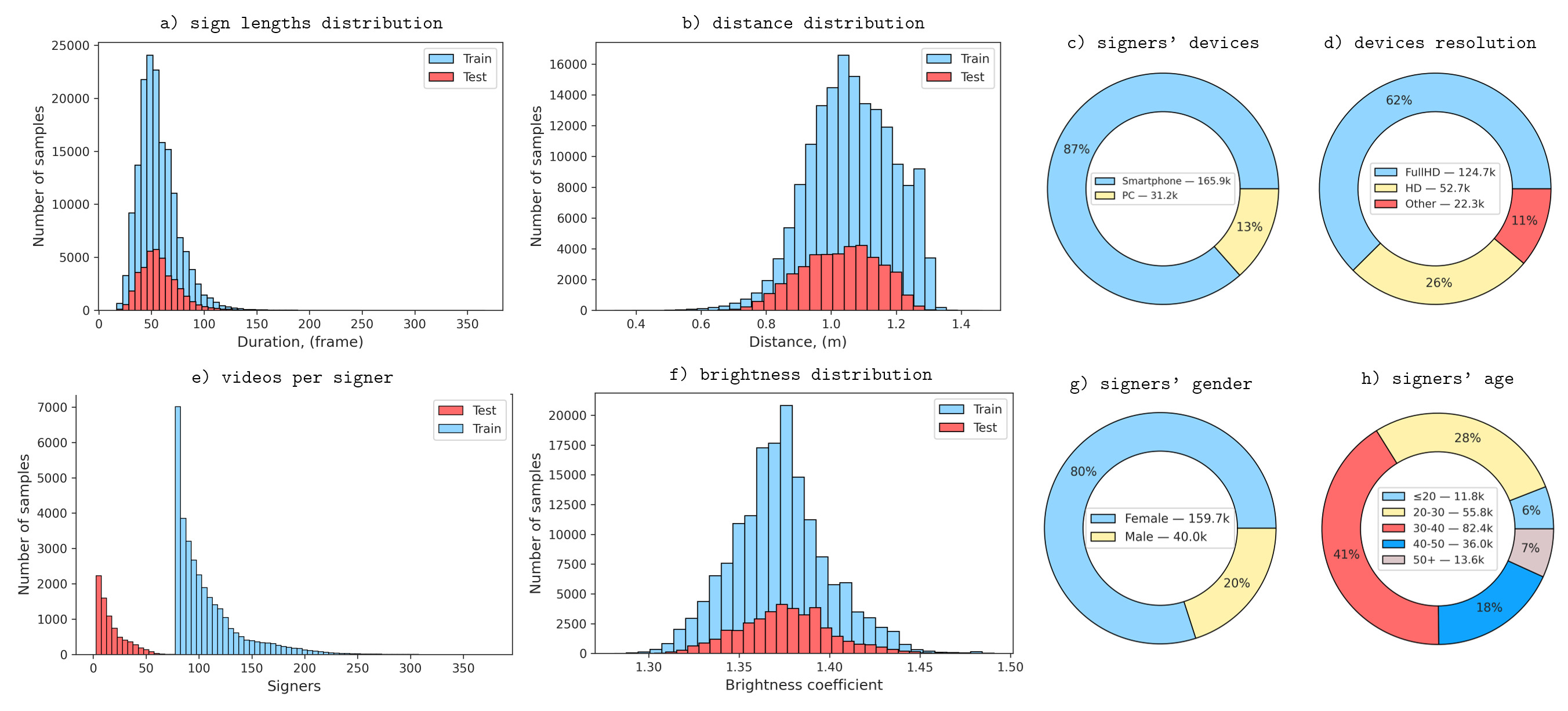}
    \caption{
    Dataset characteristics and distribution analysis.
    a) Sign length distribution. b) Distance distribution. The distance (in meters) is approximately estimated based on the length between the left and right shoulders of the signer obtained using MediaPipe~\protect\cite{mediapipe}. c) Signers' devices. d) Devices resolution. e) Number of videos per signer. f) Brightness distribution. The sample brightness is the mean pixel brightness of grayscaled video frames. g) Signers' gender; h) Signers' age. The age is determined by the MiVOLO model~\protect\cite{kuprashevich2023mivolo}.}
    \label{fig:chars}
\end{figure*}

Videos have a resolution of at least 720 pixels on the minimum side at a 30 FPS frame rate. About 62\% of videos are in FullHD format.
Distributions of some dataset characteristics are represented in Figure~\ref{fig:chars}.
Among the 381 dataset signers, 41\% are 30-40 years old, and 88\% are female.
We do not limit crowdsourcers by age and gender, and such an uneven distribution reflects the demographics of signers who wish to participate in the project.
The participants include deaf individuals, professional interpreters, sign language teachers, and CODA/SODA. Since the data collection was conducted via a crowdsourcing platform, we do not have distribution statistics for these groups. However, all participants passed an RSL proficiency test before contributing.
The dataset is divided into 80.7\% in the train and 19.3\% in the test sets.

\begin{algorithm}[!ht]
    \caption{Balanced train-test split}
    \label{alg:split}
    \begin{algorithmic}
    \State \#Notation:
    \State $\{G\} : \text{all gloss labels}$
    \State $\{S\} : \text{all signers}$
    \State $\{T\} \subset \{S\} : \text{test set signers}$
    \State $N_g : \text{samples number of gloss } g$
    \State $N_{g,s} : \text{ samples number of } g \text{ from signer } s$
    \State $R_{g,s} = N_{g,s} / N_g$
    \State \#Initialization:
    \State $p \gets 0.2$ \Comment{target test ratio}
    \State $\{T\} \gets \text{random } p \text{ from }\{S\}$ \Comment{test set signers }
    \State \#Optimization:
    \Repeat
        \State $\forall g: D_g \gets \sum_{s \in T}{R_{g,s}}$ \Comment{test gloss ratios}
        \State $d_{wst} \gets \max_g{\left|D_g - p\right|}$ \Comment{worst deviation}
        \State $g_{wst} \gets {\arg\!\max_g}{\left|D_g - p\right|}$ \Comment{worst gloss}
        \State \#Build sorted list of
        \State \#test signer candidates for replacement: 
        \If{$D_{g_{wst}}>p$} 
            \State $U \gets \text{sorted}_\text{desc }T \text{ by } R_{g_{wst}, s \in T}$
        \Else
            \State $U \gets \text{sorted}_\text{asc }T \text{ by } R_{g_{wst}, s \in T}$            
        \EndIf
        \For{$s' \text{ in } U$}
            \State \#Try to replace $s\prime$
            \State \#with signer not in test set:
            \For{$s'' \text{ in } S \setminus T$}
                \State $\forall g: D'_g \gets D_g - R_{g,s'} + R_{g,s''}$ 
                \State $d'_{wst} \gets \max_g{\left|D'_g - p\right|}$
                \If{$d'_{wst} < d_{wst}$}
                    \State \#replace the $s'$ signer in $\{T\}$
                    \State $\{T\} \gets \{T\} \cup \{s''\} \setminus \{s'\} $
                    \State break to outer Repeat
                \EndIf
            \EndFor
        \EndFor
    \Until{\text{converge}}
    \State $\text{\{test video samples\}} \gets \{\text{video: signer} \in T\}$  
    \State \textbf{Output:} \text{\{test video samples\}}

    \end{algorithmic}
\end{algorithm}

\subsection{Data Collection}
\label{app_subsec:dataset_collection}

Logos data collection pipeline follows the Slovo\cite{kapitanov2023SlovoRussianSign} pipeline in signer selection, video collection, video validation, and sign time interval annotation stages. The dataset was collected on the crowdsourcing platforms ABC Elementary and Yandex Toloka.

\paragraph{Signers Selection}
All project signers confirmed their Russian Sign Language (RSL) proficiency by passing an exam on the crowdsourcing platform ABC Elementary\footnote{\url{https://elementary.activebc.ru/}} or Yandex Toloka\footnote{\url{https://platform.toloka.ai/}}. The exam involved watching a video demonstrating an RSL sign and selecting the correct translation into Russian. Successful completion required correctly answering at least 17 out of 20 questions.

\paragraph{Video Collection.}
Signers watched a video of a correctly performed sign (video template) taken from SpreadTheSign\footnote{\url{https://spreadthesign.com/ru.ru/search/}} and then recorded a video replicating that sign. Before starting the tasks, signers reviewed the rules for video recording: 1) the gesture must match the example sign; 2) hands must remain within the frame; 3) only one person may appear in the video; 4) video does not shake; 5) video without processing; 6) video must have a short side of at least 720 pixels. Signers could record videos using a smartphone or webcam or upload pre-recorded videos from memory, watch the video, and overwrite the sign.

\paragraph{Video Validation.}  
At least three validators (SL experts: teachers, professional interpreters) who have successfully completed the validation training and exam will review the video to ensure compliance with the video recording rules. Honeypots (predefined videos with known answers to the customer) were deliberately added to the validation project to identify and eliminate unscrupulous validators.

\paragraph{Time Interval Annotation.} 
Time interval annotations were necessary to identify the start and the end of the sign dynamics and to exclude unrelated signers' movements. Each video was pre-processed to a frame rate of 30 fps, enabling the linkage of time intervals to frame numbers. All annotators have completed training and exams on time interval annotation. Honeypots were incorporated into the main tasks to identify and exclude dishonest annotators. Three annotators annotated each video, and the results were aggregated by open-source AggMe framework\footnote{\url{https://github.com/ai-forever/aggme}}.

\subsection{Train-test Split}
\label{app_subsec:split}
Given that the number of signs recorded by different signers differs, we applied a dynamic programming algorithm to find the approximate solution that has non-intersecting signers in train and test subsets and simultaneously provides approximately 20/80 test/train split ratio both for signers and for each sign samples.

The main idea is that we always select strictly 20\% signers and form the test set by the samples they demonstrated. We start with randomly selected 20\% signers. Then, in a cycle, we find a sign with a test/train balance the most different from the target 20/80 ratio and try to swap a signer from the test set and a signer not in a test set to reduce the worst disbalance,  starting from the test set signers who recorded the largest number of the scoped sign samples. So, we iteratively minimize the maximum deviation from the target test/train sign samples split ratio, keeping the test/train split ratio for signers always equal to the required value. See Algorithm~\ref{alg:split} for more details.

\section{Sign Language Recognition Pipeline}

This appendix provides additional details of the training pipeline used in the work.

\subsection{Data Pre-processing}

We randomly sample a 32-frame chain from a video with a step of 2 frames. If a sign on a video is longer than 63 frames, the sample is randomly selected within the sign duration. If extra frames are present on the video before and/or after the sign, we extend the range for sample selection up to 5 frames before and after the sign boundaries. For signs shorter than 63 frames, the sample is selected from the sign start and padded at the end by the last frame.

At the training stage \textit{Speed Up\&Slow Down} and \textit{Random Add\&Random Drop} frame sampling augmentations from Kvanchiani et al. pipeline~\shortcite{kvanchiani2024TrainingStrategiesIsolated} are applied. With probability $p=0.25$, video is accelerated twice or slowed twice with the same $p$. With $p=0.5$ we randomly drop 10\% of frames. With $p=0.25$ we truncate a sampled frame chain by 30\% and stretch it to the original size by random repeat of remaining frames.

Sampled frames are resized to 300 pix over the longest side and randomly cropped to $224 \times 224$ with square padding if needed. Frames are augmented with ColorJitter, RandomNoise, Sharpness, Flip, RandomErasing, and ImageCompression image augmentations. Augmentation parameters are set the same for every frame in a video sample.

Also we use CutMix~\cite{yun2019CutMixRegularizationStrategy} and Mixup~\cite{zhang2018MixupEmpiricalRisk} inter-sample regularization strategies.

\subsection{Training Schedule}

Training on the Logos dataset was performed on 4 Tesla H100s with 80GB RAM using batch size 16 per GPU for 50 epochs, which took about 40 hours.

For the first 5 epochs, the learning rate linearly increases from 8e-6 to 4.8e-3. Then, a cosine scheduler is used for epochs 6 to 40, reducing LR to 8e-5. Then, LR remains constant.

We use the AdamW optimizer with weight decay=0.05.

When datasets other than Logos are used, or in case of multi-dataset training, to maintain comparable training conditions, including training time, we scale training epochs number and LR schedule to keep the same number of iterations. For instance, for training on a 50\% subset of Logos, we train for 100 epochs, with cosine LR annealing from epoch 11 to epoch 80.

For tasks of transfer learning with a frozen encoder, we use a faster training schedule with reduced maximum LR: 15 total epochs with linear LR warm-up from 8e-6 to 8e-4 for 5 times shorter period, 3.5 times shorter cosine annealing from 8e-4 to 8e-5, total training iteration number -- 30\% of initial training duration. The specified number of epochs is calculated for the target dataset as described above.

\end{document}